# Calibration and Conformal Prediction for Medical Question Answering in Arabic and English: Evidence from Two Non-Parallel Benchmarks

Mariya, Stanley College of Engineering and Technology for Women, Hyderabad, India

## Abstract

Large language models (LLMs) are increasingly tested on medical exam questions, but accuracy alone does not show when an answer can be trusted. I study the calibration and uncertainty of two open-weight models, Qwen2.5-7B-Instruct and Jais-adapted-7B-chat (both run in 4-bit precision), on 2,816 English multiple-choice questions from MedMCQA and 4,927 Arabic multiple-choice questions from MedAraBench. Accuracy was 0.561 (English) and 0.430 (Arabic) for Qwen, and 0.369 and 0.308 for Jais. Raw option probabilities were overconfident in all four settings (expected calibration error 0.242 to 0.486), and a single temperature parameter removed 82% to 93% of that error. Split conformal prediction reached its nominal coverage in all 24 settings I tested (within 0.006), but the sets were large when accuracy was low: at 90% coverage, temperature-scaled sets contained on average 63% (Qwen, English), 78% (Qwen, Arabic), 86% (Jais, English) and 89% (Jais, Arabic) of the options. The margin between the top two option probabilities separated correct from incorrect answers best for Qwen on English (AUROC 0.734) and worst for Jais on Arabic (0.579). Answering only the 20% most confident questions lowered Qwen's English error from 0.439 to 0.098. Jais showed a strong prior for option C (49% of English and 60% of Arabic predictions). Scoring Latin letters only cut the Arabic share to 29% but left accuracy, raw ECE and AUROC nearly unchanged. Because the two benchmarks differ in subject, level and number of options, I do not attribute differences between languages to language itself. Code is available at https://github.com/Mariyaaleem/arabic-english-medqa-uncertainty.

## 1. Introduction

LLMs are increasingly evaluated on medical licensing and exam-style questions (Singhal et al., 2023; Jin et al., 2021; Pal et al., 2022). When a wrong answer may mislead a reader, accuracy is only part of the picture: a user also needs to know how much confidence to place in each answer. A model that is often right but states near-certainty on its errors is harder to use responsibly than one whose confidence tracks its correctness.

Calibration describes whether predicted confidence matches observed accuracy. Modern neural networks are often miscalibrated, and simple post-hoc methods such as temperature scaling can reduce the gap (Guo et al., 2017). For LLMs answering multiple-choice questions, the probability assigned to each option can be informative about correctness when the prompt format is suitable (Kadavath et al., 2022). Conformal prediction offers a complementary tool: it turns model scores into prediction sets that contain the correct answer with a user-chosen probability, under an exchangeability assumption and without further assumptions on the model (Vovk et al., 2005; Angelopoulos & Bates, 2021). It has been applied to multiple-choice question answering with LLMs (Kumar et al., 2023) and to English medical multiple-choice benchmarks, including MedMCQA and MedQA (Ke et al., 2025).

Arabic medical question answering now has dedicated benchmarks, including MedAraBench (Abu-Daoud et al., 2026) and MedArabiQ (Abu Daoud et al., 2025). These resources mainly support accuracy-based comparison of models. Calibration of Arabic LLMs has been examined at task level in the ArabicMMLU evaluation (Koto et al., 2024), and calibration across languages has been studied for multilingual models more generally (Ahuja et al., 2022; Zhou et al., 2025). For English medical multiple-choice questions, conformal prediction (Ke et al., 2025), conformal prediction combined with abstention (Machcha et al., 2026) and the relation between calibration and discrimination across clinical specialties (Testoni & Calixto, 2026) have been studied. Closest to my setting, Abouzahir et al. (2026) compared accuracy on MedAraBench with a machine-translated English version and examined model-reported confidence, which was negatively correlated with correctness; they did not measure calibration error or conformal prediction. To my knowledge, and according to a literature search in November 2026, these elements have not been combined for Arabic medical question answering.

I therefore ask four questions. RQ1: How well calibrated are open-weight LLMs on Arabic and English medical multiple-choice questions? RQ2: Does calibration differ across models and languages? RQ3: Do split conformal prediction sets reach their nominal coverage, and how informative are they? RQ4: Does selective prediction, answering only when confident, reduce errors? My contributions are: (i) a reproducible protocol for measuring calibration, conformal coverage and selective prediction from option probabilities; (ii) results for two models on two benchmarks, with confidence intervals; (iii) evidence that a low calibration error after temperature scaling can coexist with confidence that barely separates correct from incorrect answers, and that conformal set size tracks accuracy; and (iv) open code at https://github.com/Mariyaaleem/arabic-english-medqa-uncertainty.

# 2. Related Work

## 2.1 Calibration of neural networks and LLMs

Guo et al. (2017) showed that modern deep networks tend to be miscalibrated and that temperature scaling is a simple and effective correction. Kadavath et al. (2022) examined whether language models know what they know and reported that larger models are well calibrated on multiple-choice and true/false questions in suitable formats. Selective classification, in which a model abstains when uncertain, was extended to deep neural networks by Geifman and El-Yaniv (2017), who used a confidence threshold to guarantee a target error rate with high probability. Testoni and Calixto (2026) compared calibration (ECE) with discrimination (AUROC) for uncertainty estimates of ten open-source LLMs on two English clinical multiple-choice datasets and found that reliability varies with clinical specialty and question type.

## 2.2 Conformal prediction

Conformal prediction provides distribution-free coverage guarantees under exchangeability (Vovk et al., 2005), and Angelopoulos and Bates (2021) give an accessible introduction. Kumar et al. (2023) applied conformal prediction to multiple-choice question answering with LLMs, and Quach et al. (2024) extended the idea to open-ended generation. Ke et al. (2025) applied conformal prediction to medical multiple-choice questions from MedMCQA, MedQA and MMLU, using small Llama 3.2 and Qwen2.5 models (1 to 3 billion parameters). Machcha et al. (2026) combined conformal prediction with abstention options and perturbed questions for medical multiple-choice questions. Angelopoulos et al. (2024) generalized coverage to broader risk controls.

## 2.3 Arabic and medical benchmarks

MedAraBench is a large Arabic medical question answering benchmark with 24,883 multiple-choice questions across 19 specialties and five levels corresponding to years of medical study (Abu-Daoud et al., 2026); its authors report a training split of 19,894 and a test split of 4,989 questions. An earlier, smaller resource is MedArabiQ (Abu Daoud et al., 2025), where Jais-13b-chat performed poorly on the closed-format tasks. Abouzahir et al. (2026) evaluated larger open models on MedAraBench and on a machine-translated English version of it and reported lower accuracy in Arabic for most models; their confidence analysis used model-reported confidence and not token probabilities. ArabicMMLU (Koto et al., 2024) benchmarks Arabic school-exam knowledge and scores models by comparing next-token probabilities of option letters, in Latin or Arabic script, and the Jais models are Arabic-centric open LLMs (Sengupta et al., 2023; Inception, 2024). For English, MedQA (Jin et al., 2021) and MedMCQA (Pal et al., 2022) are widely used. Calibration has also been studied across languages. Ahuja et al. (2022) found clear miscalibration of massively multilingual models in low-resource languages and in languages typologically distant from English, and Zhou et al. (2025) reported that non-English languages are calibrated worse than English across several model families. For Arabic, Koto et al. (2024) analyzed calibration at task level on ArabicMMLU and found confidence and accuracy to be strongly correlated across the 40 tasks ($r > 0.9$) for BLOOMZ, AceGPT-chat and Jais-chat.

# 3. Methods

## 3.1 Benchmarks

English. I used the validation split of MedMCQA (Pal et al., 2022), which contains 4,183 questions with four options each and a labeled correct answer. I kept the 2,816 questions marked as single-choice, so that chance accuracy is 0.250. The correct answer is not uniformly distributed over positions (A 33.7%, B 26.1%, C 21.9%, D 18.3%), so a strategy that always answers A scores 0.337. The largest subjects are Dental (845 questions), Pathology (259), Surgery (249) and Medicine (185).

Arabic. I used the test split of MedAraBench (Abu-Daoud et al., 2026). The released file I used has 4,959 rows, whereas the benchmark paper reports 4,989 test questions; I report results on the file as released. I removed 32 rows (1 empty question, 21 whose answer was not an option letter, and 10 whose answer pointed to an empty option), leaving 4,927 questions. Questions have 4 options (3,966 questions), 5 options (959) or 6 options (2), so chance accuracy is 0.240. Correct answers are spread evenly over positions A to D (about 0.24 each); the benchmark's authors balanced answer positions (Abu-Daoud et al., 2026, Appendix F). The largest subjects are Anatomy (1,180 questions), Chemistry and Physics (781), Physiology (667), Cell and Molecular Biology (630), Biochemistry (365) and Statistics (361). The benchmark's five levels correspond to years of medical study, and its authors note that the difficulty distribution is skewed toward easier levels (Abu-Daoud et al., 2026, Section 6). The dataset license is not stated, so I did not redistribute any question text; my released files contain no question text.

The two benchmarks differ in source exams, subject mix, apparent stage of training and number of options. They are not parallel translations of each other (a machine-translated English version of MedAraBench was used by Abouzahir et al., 2026; I did not use it). Differences between the English and Arabic results are therefore confounded with differences between datasets (Section 3.7).

## 3.2 Models and inference

I evaluated Qwen2.5-7B-Instruct (Qwen et al., 2024) and Jais-adapted-7B-chat (Inception, 2024), an Arabic-extended model built on Llama 2 (Touvron et al., 2023) (Hugging Face repository inception42/jais-adapted-7b-chat, revision f2de64b0). Both were loaded in 4-bit NF4 quantization with float16 computation using the bitsandbytes library on a single NVIDIA T4 GPU (15 GB), with fixed seed 42. Each question required one forward pass; no text was generated.

## 3.3 Prompts and option probabilities

Each prompt contained a fixed English instruction (answer the medical multiple-choice question with the letter of the correct option only), the question, and the options labeled A to F. For Qwen I used the model's chat template. For Jais I used the model-specific template with an instruction block, a human turn and an AI turn, ending with a response marker, because Jais is not trained on the Qwen format. The instruction block used a one-sentence system line, a shortened version of the longer system prompt on the Jais family model cards. In both cases the instruction was English and the question and options were in the language of the benchmark.

I read the model's next-token log-probabilities for the option letters and applied a softmax restricted to the valid options, giving a probability vector p over the K options. The predicted answer is the option with the largest probability, and the confidence is that probability. As a format check I report the mass on letters, the total next-token probability that falls on valid option tokens before renormalization. It was 1.000 for Qwen in both languages.

Two implementation choices were made on small development samples and are disclosed here. First, for Qwen I compared three prompt endings on 50 English questions; two reached a mass on letters of 1.000 and I kept the simpler one (an earlier ending with a trailing space after the answer cue had given 0.888, and a third ending gave 0.263); I selected by format compliance, not by accuracy. Second, Jais frequently answered with the Arabic spelling of an option letter. On a 50-question Arabic sample, counting only Latin letters gave a mean mass of 0.497, and a diagnostic over 20 questions showed that Arabic letters (for example the letter used for option C) held about 42% of the next-token probability. For Jais I therefore added the probability of the Arabic letter to that of the Latin letter for each option (A and the Arabic letter alif, B and baa, C and jeem, D and dal, and so on). This raised the mean mass to 0.926 on Arabic and left English unchanged (0.899; Arabic letters held almost no probability on English prompts, and the remaining mass went mainly to the word 'The'). I applied the same scoring to Jais in both languages. Comparing option letters written in Latin or Arabic script follows the evaluation practice of ArabicMMLU (Koto et al., 2024). A run that scored Latin letters only on English gave the same predicted option for all 2,816 questions and the same accuracy, ECE and AUROC to three decimals (Appendix C, Table C1). On all 4,927 Arabic questions, Latin-only scoring lowered the mean mass on letters from 0.926 to 0.476 and changed the predicted option for 1,746 questions (35.4%); accuracy (0.297 against 0.308), raw ECE (0.310 against 0.309) and AUROC (0.592 against 0.579) changed little, but the share of C predictions fell from 59.6% to 28.5% (Appendix C, Table C1).

## 3.4 Metrics

Accuracy is the share of questions for which the predicted option equals the labeled answer. Expected calibration error (ECE; Guo et al., 2017) partitions predictions into M = 15 equal-width confidence bins B_m and averages the gap between accuracy and mean confidence, weighted by bin size:

*ECE = sum over m = 1..M of (|B_m| / n) x |acc(B_m) - conf(B_m)|*

The Brier score is the sum over options of the squared difference between the predicted probability and the one-hot label, averaged over questions; it ranges from 0 to 2, and a uniform guess over K options scores 1 - 1/K (0.75 for four options). To measure how well confidence ranks correct above incorrect answers, I report the area under the ROC curve (AUROC) of the margin between the top two option log-scores as a predictor of correctness. All confidence intervals are 95% percentile bootstrap intervals over questions with 500 resamples (Efron & Tibshirani, 1993).

### 3.5 Temperature scaling

I divided the option log-scores by a single temperature T, fitted on a calibration subset by minimizing negative log-likelihood. To estimate performance honestly I repeated 20 random 50/50 splits (seed 42), fitting T on one half and computing ECE on the other, and report means and standard deviations.

### 3.6 Split conformal prediction and selective prediction

For each question i with true option y_i, the nonconformity score is s_i = 1 - p_i,y. For a target error level alpha, I took the ceil((m + 1)(1 - alpha))-th smallest of m calibration scores as the threshold q, and the prediction set for a new question contains every option k with 1 - p_k <= q. If calibration and test questions are exchangeable, the set contains the correct option with probability at least 1 - alpha (Vovk et al., 2005; Angelopoulos & Bates, 2021); the guarantee is marginal, not conditional on subject or set size.

I repeated 100 random splits, each using 25% of the questions to fit the temperature, 25% for conformal calibration and 50% for testing, for alpha in {0.05, 0.10, 0.20}, with raw probabilities (T = 1) and temperature-scaled probabilities. I report empirical coverage, mean set size (also as a share of the number of options), the share of singleton sets and the share of sets containing every option. For selective prediction (Geifman & El-Yaniv, 2017), I ranked questions by the margin between the top two log-scores, answered the most confident fraction c, and report the error rate on the answered questions (risk) at c = 0.2, 0.4, 0.6 and 0.8, and the area under the risk-coverage curve (AURC).

### 3.7 Comparing languages and models

Because the benchmarks are not parallel, I (i) report each benchmark's chance level, (ii) emphasize calibration, coverage and set size relative to the number of options and not raw accuracy, and (iii) do not attribute any difference to language alone. Both models were run on the same questions, so differences between models within a benchmark are not confounded by the data, but they are confounded by model-specific prompt templates and scoring (Section 3.3).

## 4. Experimental Setup

| Setting | Value |
|---|---|
| Hardware | NVIDIA Tesla T4 (about 15 GB), Google Colab |
| Models | Qwen2.5-7B-Instruct; Jais-adapted-7B-chat (4-bit NF4, float16 compute) |
| English data | MedMCQA validation, single-choice only: 2,816 questions, 4 options, chance 0.250 |
| Arabic data | MedAraBench test: 4,927 usable questions, 4 to 6 options, chance 0.240 |
| Calibration metrics | ECE with 15 bins; Brier (sum over options); AUROC of top-two margin |
| Temperature scaling | 20 random 50/50 splits, seed 42 |
| Conformal prediction | 100 random 25/25/50 splits; alpha 0.05, 0.10, 0.20; raw and scaled probabilities |
| Confidence intervals | 95% percentile bootstrap over questions, 500 resamples |
| Software | Python 3.13; PyTorch 2.11; Transformers 5.22; Accelerate 1.15; bitsandbytes 0.50.2; scikit-learn 1.6; SciPy 1.16; NumPy 2.4; pandas 2.2 (Google Colab runtime, November 2026) |
| Code | https://github.com/Mariyaaleem/arabic-english-medqa-uncertainty |

*Table 1. Experimental settings.*

# 5. Results

## 5.1 Accuracy and calibration

Qwen was more accurate than Jais in both languages, and both models were more accurate on English than on Arabic (Table 2); all accuracies are above chance, but Jais on Arabic is only about seven points above it. Raw probabilities were overconfident in every setting. Qwen's probabilities were close to saturated: it gave confidence above 0.99 to 68.9% of English questions, of which 65.6% were answered correctly, and to 54.6% of Arabic questions, of which 52.0% were answered correctly. Jais never exceeded 0.99, with mean confidence near 0.61 in both languages. The raw Brier scores of all four settings are worse than the uniform-guess baseline (Table 3).

| Model, language | n | Chance | Accuracy (95% CI) | Mass on letters | Mean conf. | AUROC (95% CI) |
|---|---|---|---|---|---|---|
| Qwen, English | 2,816 | 0.250 | 0.561 (0.541 to 0.582) | 1.000 | 0.948 | 0.734 (0.714 to 0.752) |
| Qwen, Arabic | 4,927 | 0.240 | 0.430 (0.417 to 0.444) | 1.000 | 0.916 | 0.646 (0.632 to 0.660) |
| Jais, English | 2,816 | 0.250 | 0.369 (0.352 to 0.386) | 0.899 | 0.611 | 0.600 (0.579 to 0.623) |
| Jais, Arabic | 4,927 | 0.240 | 0.308 (0.296 to 0.321) | 0.926 | 0.618 | 0.579 (0.562 to 0.596) |

*Table 2. Accuracy, format compliance, mean confidence and AUROC of the top-two margin for correctness.*

A single temperature removed most of the calibration error (Table 3). The fitted temperature was larger for Qwen (8.7 and 10.9) than for Jais (3.4 and 5.7), reflecting Qwen's saturated probabilities, and for both models it was larger on Arabic than on English. After scaling, ECE was between 0.035 and 0.068, a reduction of 82% to 93%. The reliability diagrams (Figure 1) show raw curves far below the diagonal and scaled curves close to it except in the sparse top bins (Table B1); for Qwen the scaled confidences spread across 0.27 to 0.83, whereas for Jais 98% of Arabic and 77% of English test questions fall into two bins centred near 0.27 to 0.35, close to chance level.

| Model, language | ECE raw (95% CI) | ECE raw, 20 splits | ECE scaled, 20 splits | T (mean, SD) | Brier (uniform guess) |
|---|---|---|---|---|---|
| Qwen, English | 0.386 (0.367 to 0.407) | 0.388 | 0.068 | 8.71, 0.23 | 0.806 (0.750) |
| Qwen, Arabic | 0.486 (0.472 to 0.500) | 0.487 | 0.035 | 10.90, 0.34 | 1.025 (0.760) |
| Jais, English | 0.242 (0.226 to 0.260) | 0.244 | 0.042 | 3.40, 0.17 | 0.815 (0.750) |
| Jais, Arabic | 0.309 (0.297 to 0.322) | 0.308 | 0.042 | 5.70, 0.36 | 0.899 (0.760) |

*Table 3. Calibration before and after temperature scaling. The 20-split columns use the same random splits, so raw and scaled values are directly comparable.*

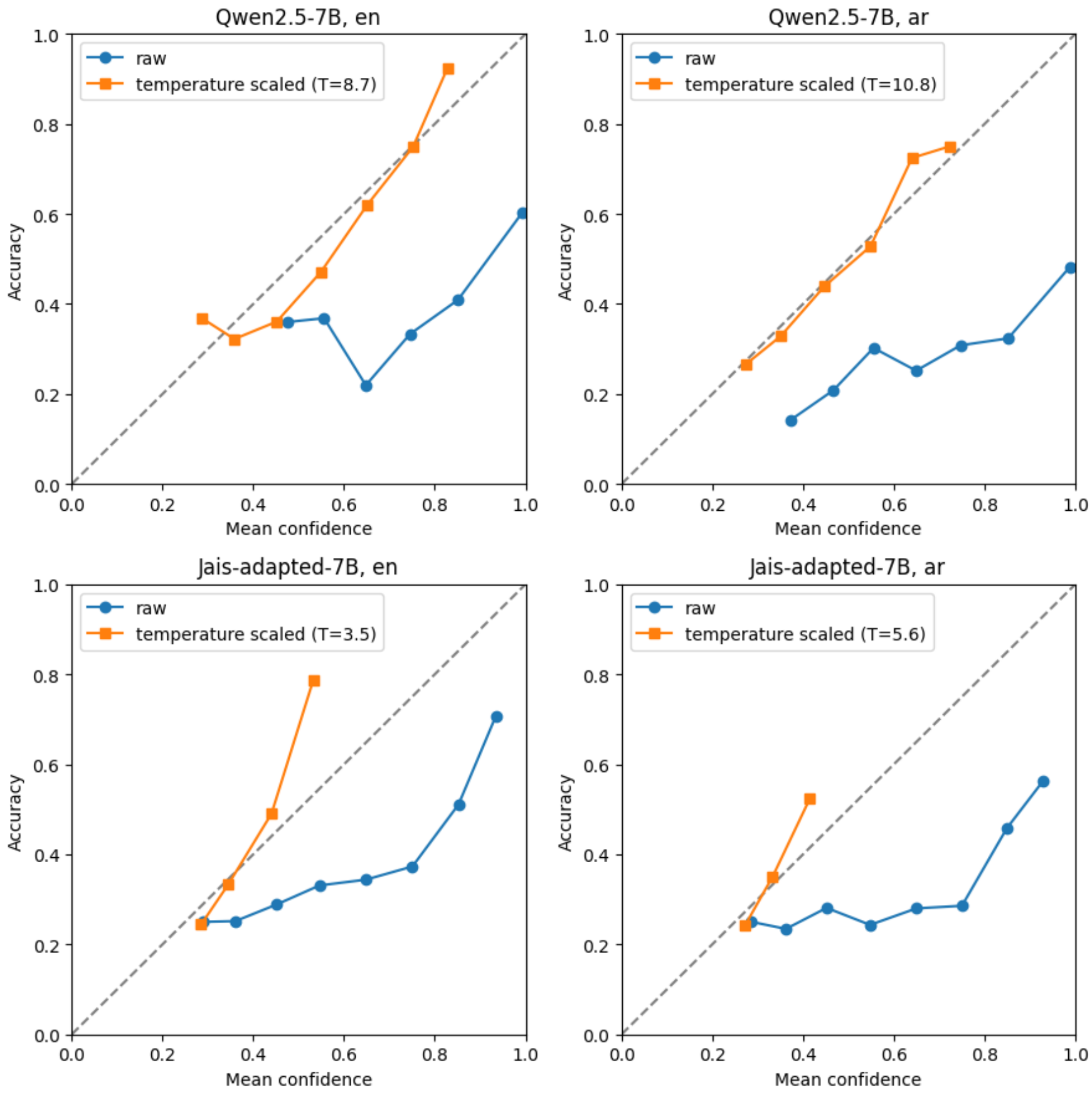

*Figure 1. Reliability diagrams, raw and temperature scaled, for each model and language (one random 50/50 split; the temperatures in the legends come from that split and differ slightly from the 20-split means in Table 3). Questions per bin for the scaled curves are given in Table B1.*

Low calibration error after scaling should not be read as high usefulness. For Jais, scaled confidence is close to the base rate for almost every question, which makes ECE small while the AUROC remains between 0.579 and 0.600. For Qwen, the scaled confidences are informative and the curve follows the diagonal over a wide range.

## 5.2 Conformal prediction

Empirical coverage was within 0.006 of the target in all 24 settings (two models, two languages, raw and scaled probabilities, three error levels). Informativeness, however, followed accuracy (Table 4). At 90% coverage, Qwen's scaled sets contained on average 63% of the options in English and 78% in Arabic, and Jais's contained 86% and 89%. A random subset of the options would need 90% of them to reach the same coverage. Many questions received a set containing every option: 25% and 40% for Qwen, and 55% and 67% for Jais (scaled probabilities, English and Arabic). Singleton sets were 23% of English questions for Qwen but only 1% for Jais in English and 0.2% in Arabic. Raw and scaled probabilities gave similar mean set sizes; scaling gave fewer singleton sets and, for Qwen, fewer all-option sets. Bootstrap intervals for coverage and set size are in Appendix C (Table C2); the intervals for Qwen on Arabic and Jais on English, and for Jais on English and Arabic, partly overlap, so those differences should not be over-read.

| Model, language | Probabilities | Coverage (target 0.90) | Set size, options | Set size / K | Singleton | All options |
|---|---|---|---|---|---|---|
| Qwen, English | raw | 0.905 | 2.51 | 0.629 | 0.303 | 0.331 |
| | scaled | 0.905 | 2.51 | 0.627 | 0.234 | 0.246 |

| Model, language | Probabilities | Coverage (target 0.90) | Set size, options | Set size / K | Singleton | All options |
|---|---|---|---|---|---|---|
| Qwen, Arabic | raw | 0.902 | 3.25 | 0.773 | 0.130 | 0.517 |
| | scaled | 0.901 | 3.26 | 0.779 | 0.049 | 0.396 |
| Jais, English | raw | 0.902 | 3.36 | 0.841 | 0.058 | 0.591 |
| | scaled | 0.902 | 3.44 | 0.859 | 0.011 | 0.553 |
| Jais, Arabic | raw | 0.901 | 3.63 | 0.867 | 0.019 | 0.589 |
| | scaled | 0.900 | 3.68 | 0.890 | 0.002 | 0.666 |

*Table 4. Split conformal prediction at alpha = 0.10, averaged over 100 random splits. Singleton and all-options columns are shares of test questions. Results for alpha = 0.05 and 0.20 are in Appendix A (Tables A1 and A2).*

The coverage guarantee is marginal. In a separate analysis of Qwen at alpha = 0.10, one-option sets were correct for 83.8% (raw) and 86.7% (scaled) of English questions, and for 70.8% (raw) and 78.1% (scaled) of Arabic questions, while sets containing all options trivially covered every question. Coverage within subjects varied around the target for temperature-scaled Qwen: from 0.81 to 0.96 across English subjects with at least 30 questions, and from 0.78 to 0.96 across Arabic subjects, with the largest shortfalls for Microbiology in English (0.812, n = 74) and Surgery in Arabic (0.783, n = 128). These per-subject figures are exploratory.

## 5.3 Selective prediction

Abstaining on the least confident questions helped, but by very different amounts (Table 5). For Qwen on English, answering only the 20% most confident questions reduced the error rate from 0.439 to 0.098. In Arabic the same procedure gave 0.354 against 0.570. For Jais the gains were small: 0.448 against 0.631 in English and 0.543 against 0.692 in Arabic. The AURC ordering matches the AUROC ordering, and for Qwen the raw max-probability confidence gave essentially the same curves as the margin (AURC 0.248 against 0.249 in English). For Jais it gave somewhat lower risk (AURC 0.520 against 0.529 in English and 0.607 against 0.616 in Arabic). The AUROC of the maximum probability was also slightly higher than that of the margin in all four settings, and the ordering of the settings is the same under both rankings (Appendix C, Table C3).

| Model, language | Overall error | Risk at 80% | Risk at 60% | Risk at 40% | Risk at 20% | AURC |
|---|---|---|---|---|---|---|
| Qwen, English | 0.439 | 0.386 | 0.308 | 0.217 | 0.098 | 0.249 |
| Qwen, Arabic | 0.570 | 0.537 | 0.493 | 0.443 | 0.354 | 0.440 |
| Jais, English | 0.631 | 0.615 | 0.590 | 0.551 | 0.448 | 0.529 |
| Jais, Arabic | 0.692 | 0.683 | 0.658 | 0.634 | 0.543 | 0.616 |

*Table 5. Risk (error rate on answered questions) when answering the most confident 80%, 60%, 40% and 20% of questions, ranked by the top-two margin.*

## 5.4 Answer-position bias

Qwen's predicted letters followed the true distribution closely. Jais showed a strong prior for option C: it predicted C for 49.0% of English questions (true share 21.9%) and 59.6% of Arabic questions (true share 23.7%), and predicted A for only 13.0% and 4.4% (Table 6). Always answering C on the Arabic set would give 23.7% accuracy, against 30.8% for Jais under union scoring, so under that scoring a large part of its accuracy is accounted for by this prior. The size of the prior depends on the letter scoring: with Latin letters only, Jais predicted C for 28.5% of the Arabic questions and its accuracy was 0.297 (Appendix C, Table C1). Accuracy was lower on questions with low mass on letters in English (0.315 against 0.411), with little difference in Arabic (0.292 against 0.313).

| Benchmark | Source | A | B | C | D | E |
|---|---|---|---|---|---|---|
| English | True | 0.337 | 0.261 | 0.219 | 0.183 | |
| English | Qwen predicted | 0.319 | 0.262 | 0.258 | 0.161 | |
| English | Jais predicted | 0.130 | 0.216 | 0.490 | 0.164 | |
| Arabic | True | 0.239 | 0.240 | 0.237 | 0.236 | 0.047 |
| Arabic | Qwen predicted | 0.226 | 0.259 | 0.242 | 0.247 | 0.026 |
| Arabic | Jais predicted | 0.044 | 0.112 | 0.596 | 0.239 | 0.009 |

*Table 6. True and predicted answer-letter shares on the full runs (option E exists only for the 5- and 6-option Arabic questions).*

# 6. Discussion

RQ1 and RQ2. Both models were overconfident in both languages, but to very different degrees, and a one-parameter temperature fixed most of the calibration error in every case. For both models, Arabic needed a higher temperature and showed higher raw ECE than English. This pattern is consistent across two models, but it cannot be attributed to language, because the benchmarks differ in subject, level and option count.

RQ3. Split conformal prediction delivered its coverage guarantee in every setting, which makes it a trustworthy wrapper around any of these models. It does not make the model more informative; Kumar et al. (2023) likewise found that conformal uncertainty was tightly correlated with accuracy for LLaMA-13B on multiple-choice questions. When accuracy is low, the sets that meet the guarantee grow to include most or all of the options, and a large share of questions receive an uninformative set. The guarantee is also marginal: for Qwen, singleton sets were correct less often than the nominal level, and coverage varied across subjects.

RQ4. Selective prediction depends on how well confidence ranks correct above incorrect answers. It worked well for Qwen on English and poorly for Jais, whose confidence separated correct from incorrect answers poorly and reflected a strong letter prior. Practitioners should therefore report discrimination (AUROC or risk-coverage) alongside calibration error, because a model can have a small ECE after scaling simply by being uncertain about everything.

## 6.1 Limitations

- The English and Arabic benchmarks are not parallel, and they differ in subject, apparent level and number of options, so language effects are confounded with dataset effects. Evaluating both models on a machine-translated English version of MedAraBench (Abouzahir et al., 2026) would remove this confound and is a natural next step.
- The study uses exam-style multiple-choice questions. It is not clinical validation and does not support claims about clinical decision support.
- Only two 7B-scale models were evaluated, in 4-bit precision, with one prompt per model. Results may change with other prompts, models or precision. Jais used a model-specific template that I shortened, and its mass on letters was only 0.899 (English) and 0.926 (Arabic). Scoring Arabic questions with Latin letters only changed the predicted option for 35% of them and cut the share of C predictions from 59.6% to 28.5%, although accuracy, raw ECE and AUROC changed little (Appendix C, Table C1).

- Jais showed a strong letter-C prior under the scoring used for the main results. Because Latin-only scoring halved the Arabic C share while accuracy, raw ECE and AUROC changed little, the weak Jais results are not explained by that prior alone. They should not be read as a measure of Arabic ability or of the model in general. I did not test prior-correction methods.
- The development choices in Section 3.3 were made on small samples (50 questions) and evaluated on the same benchmarks, although the selection criterion was format compliance and not accuracy.
- Temperature scaling uses a single parameter, and the scaled reliability curves lie above the diagonal in the top bins, so the models are underconfident there (for example accuracy 0.787 against mean confidence 0.532 for Jais in English), and those bins hold few questions (28 to 61 in three of the four runs; Table B1). Conformal guarantees rely on exchangeability and hold marginally, not for every subject or set size. Repeated random splits overlap, so the reported standard deviations describe split variability and not sampling uncertainty about the population.
- The author's own Arabic comprehension is limited; no fluent Arabic reader has checked the Arabic questions or the model errors, so problems in the benchmark text, or in my reading of the Arabic results, may remain.

### 6.2 Responsible use

The models and methods studied here are not intended for patient care. Calibration and coverage results on exam-style questions should not be read as evidence that a model is safe for clinical use.

## 7. Conclusion

On English and Arabic medical multiple-choice questions, two open-weight LLMs were strongly overconfident, and a single temperature parameter removed most of the calibration error. Split conformal prediction met its coverage guarantee in every setting, but its sets were informative only when the underlying model was accurate, and selective prediction helped substantially only for the most accurate model. A low calibration error therefore does not show that a model's confidence is useful, and I recommend reporting discrimination and set size alongside calibration. Extending the study to parallel bilingual benchmarks, more models and prior-corrected scoring is a natural next step.

## Declarations

AI assistance: The main results presented in this paper were obtained by the author. I used a generative AI assistant to help plan the study, debug analysis code, review statistical and mathematical arguments, check references, and improve the text. The author takes responsibility for the final content of this work.

Acknowledgments: I thank Nadia Anjum for guidance and Mohammed Izrar uddin for help during this work. Any errors are my own.

Data and code availability: MedMCQA is publicly available (Apache License 2.0 according to its Hugging Face dataset card; Pal et al., 2022). MedAraBench is released by its authors (https://github.com/nyuad-cai/MedAraBench, as given by Abu-Daoud et al., 2026; the repository README states no license); I do not redistribute it. Code and summary result tables are at https://github.com/Mariyaaleem/arabic-english-medqa-uncertainty.

# Appendix A. Conformal prediction at other target levels

Table A1 and Table A2 give split conformal results at the two other target levels, averaged over the same 100 random splits as Table 4. They are taken from summary_conformal.csv in the repository. Coverage stayed close to the target for both models, languages and probability variants, and set sizes grew with the target coverage.

| Model, language | Probabilities | Coverage (target 0.95) | Set size, options | Set size / K | Singleton | All options |
|---|---|---|---|---|---|---|
| Qwen, English | raw | 0.954 | 2.99 | 0.748 | 0.203 | 0.532 |
| | scaled | 0.953 | 2.96 | 0.740 | 0.152 | 0.445 |
| Qwen, Arabic | raw | 0.952 | 3.65 | 0.868 | 0.067 | 0.700 |
| | scaled | 0.951 | 3.67 | 0.877 | 0.017 | 0.624 |
| Jais, English | raw | 0.952 | 3.66 | 0.915 | 0.026 | 0.770 |
| | scaled | 0.952 | 3.70 | 0.925 | 0.003 | 0.742 |
| Jais, Arabic | raw | 0.952 | 3.90 | 0.930 | 0.006 | 0.766 |
| | scaled | 0.951 | 3.95 | 0.949 | 0.000 | 0.802 |

*Table A1. Split conformal prediction at alpha = 0.05 (target coverage 0.95), averaged over 100 random splits. Singleton and all-options columns are shares of test questions.*

| Model, language | Probabilities | Coverage (target 0.80) | Set size, options | Set size / K | Singleton | All options |
|---|---|---|---|---|---|---|
| Qwen, English | raw | 0.802 | 1.88 | 0.470 | 0.482 | 0.112 |
| | scaled | 0.802 | 1.85 | 0.463 | 0.414 | 0.044 |
| Qwen, Arabic | raw | 0.802 | 2.62 | 0.622 | 0.259 | 0.262 |
| | scaled | 0.803 | 2.61 | 0.627 | 0.133 | 0.155 |
| Jais, English | raw | 0.804 | 2.85 | 0.711 | 0.126 | 0.314 |
| | scaled | 0.804 | 2.90 | 0.724 | 0.053 | 0.246 |
| Jais, Arabic | raw | 0.800 | 3.14 | 0.750 | 0.074 | 0.325 |

| Model, language | Probabilities | Coverage (target 0.80) | Set size, options | Set size / K | Singleton | All options |
|---|---|---|---|---|---|---|
| | scaled | 0.801 | 3.16 | 0.769 | 0.033 | 0.376 |

*Table A2. Split conformal prediction at alpha = 0.20 (target coverage 0.80), averaged over 100 random splits. Singleton and all-options columns are shares of test questions.*

## Appendix B. Bins of the temperature-scaled reliability diagrams

Table B1 gives the bins behind the temperature-scaled curves in Figure 1. The diagrams use one random 50/50 split, so the test split holds 1,408 English and 2,464 Arabic questions. Bins with no questions are omitted.

| Model, language | Mean confidence | Accuracy | Questions in bin | Share of test split |
|---|---|---|---|---|
| Qwen, English | 0.287 | 0.368 | 19 | 1.3% |
| | 0.358 | 0.322 | 180 | 12.8% |
| | 0.452 | 0.361 | 277 | 19.7% |
| | 0.549 | 0.471 | 244 | 17.3% |
| | 0.652 | 0.621 | 248 | 17.6% |
| | 0.752 | 0.749 | 271 | 19.2% |
| | 0.828 | 0.923 | 169 | 12.0% |
| Qwen, Arabic | 0.275 | 0.267 | 210 | 8.5% |
| | 0.351 | 0.329 | 840 | 34.1% |
| | 0.447 | 0.439 | 684 | 27.8% |
| | 0.549 | 0.528 | 477 | 19.4% |
| | 0.640 | 0.724 | 225 | 9.1% |
| | 0.723 | 0.750 | 28 | 1.1% |
| Jais, English | 0.285 | 0.244 | 213 | 15.1% |
| | 0.345 | 0.333 | 869 | 61.7% |
| | 0.440 | 0.491 | 265 | 18.8% |
| | 0.532 | 0.787 | 61 | 4.3% |
| Jais, Arabic | 0.273 | 0.243 | 1054 | 42.8% |
| | 0.333 | 0.350 | 1368 | 55.5% |
| | 0.415 | 0.524 | 42 | 1.7% |

*Table B1. Temperature-scaled reliability bins (mean confidence, accuracy and number of questions) for the split used in Figure 1. For Jais, two bins hold 98% of the Arabic and 77% of the English test questions.*

## Appendix C. Robustness and uncertainty checks

The checks below use the saved option scores and were run with a separate script, whose random splits differ from those of the main analysis. It reproduced accuracy, raw ECE and margin AUROC of Tables 2 and 3 to three decimals; scaled ECE agreed within 0.006, fitted temperatures within 0.25 and mean set sizes within 0.04, so point estimates in this appendix can differ slightly from Tables 3 and 4.

## C.1 Latin-only versus union scoring of option letters (Jais)

Table C1 compares scoring the option letters in Latin script only with adding the probability of the Arabic letter spellings (union scoring, used for all main results). On English, Arabic letters held almost no probability and both scorings gave the same predicted option for all 2,816 questions. On Arabic, union scoring raised the mean mass on letters from 0.476 to 0.926 and changed the predicted option for 1,746 of 4,927 questions (35.4%). Accuracy, raw ECE and the AUROC of the margin were close under both scorings (accuracy lower by 0.011 and AUROC higher by 0.013 with Latin letters only), but the share of option C predictions was 28.5% with Latin letters only against 59.6% with union scoring. Paired bootstrap intervals (2,000 resamples over questions) for Latin only minus union were −0.011 (−0.022 to 0.001) for accuracy, 0.013 (−0.007 to 0.035) for the AUROC of the margin and 0.000 (−0.011 to 0.013) for raw ECE. An exact McNemar test on accuracy gave $p = 0.065$ (385 questions correct only with Latin letters, 439 only with union scoring). None of the three differences is distinguishable from zero at the 5% level, although the accuracy difference is borderline.

| Metric | English, Latin only | English, union | Arabic, Latin only | Arabic, union |
|---|---|---|---|---|
| Accuracy | 0.369 | 0.369 | 0.297 | 0.308 |
| Mean confidence | 0.611 | 0.611 | 0.607 | 0.618 |
| Mean mass on letters | 0.899 | 0.899 | 0.476 | 0.926 |
| ECE, raw | 0.242 | 0.242 | 0.310 | 0.309 |
| AUROC of top-two margin | 0.600 | 0.600 | 0.592 | 0.579 |
| Share of predictions that are option C | 0.490 | 0.490 | 0.285 | 0.596 |
| Questions with a different predicted option | 0 of 2,816 | (reference) | 1,746 of 4,927 | (reference) |

*Table C1. Jais on English (n = 2,816) and Arabic (n = 4,927): Latin-only scoring versus union scoring (used for the main results).*

## C.2 Intervals for conformal coverage and set size

Table C2 gives 95% percentile intervals from 500 bootstrap replicates for temperature-scaled conformal prediction at alpha = 0.10. Each replicate draws a random 25/25/50 split and resamples the questions within each part, so the intervals describe what one calibration split would give and are wider than the uncertainty of the 100-split means in Table 4. The intervals for set size / K of Qwen on English and Qwen on Arabic do not overlap; those of Qwen on Arabic and Jais on English, and of Jais on English and Arabic, partly overlap.

| Model, language | Coverage (target 0.90) | Set size, options | Set size / K |
|---|---|---|---|
| Qwen, English | 0.860 to 0.935 | 2.24 to 2.71 | 0.560 to 0.677 |
| Qwen, Arabic | 0.867 to 0.925 | 3.06 to 3.43 | 0.732 to 0.819 |
| Jais, English | 0.861 to 0.935 | 3.23 to 3.57 | 0.807 to 0.892 |
| Jais, Arabic | 0.869 to 0.931 | 3.57 to 3.80 | 0.862 to 0.919 |

*Table C2. 95% bootstrap intervals (500 replicates) for coverage and set size of split conformal prediction with temperature-scaled probabilities at alpha = 0.10.*

## C.3 AUROC and AURC of the top-two margin and of the maximum probability

Table C3 compares the AUROC of the top-two margin, used in the main text, with that of the maximum option probability. The maximum probability was slightly higher in all four settings (by 0.002 for Qwen and by 0.015 to 0.016 for Jais), and the ordering of the settings is the same. For the risk-coverage curves, the AURC of the maximum probability was within 0.002 of the margin for Qwen and lower by about 0.009 for Jais (0.520 against 0.529 in English and 0.607 against 0.616 in Arabic); the risk when answering the 20% most confident questions was also lower for Jais (0.430 against 0.448 in English and 0.531 against 0.543 in Arabic).

| **Model, language** | **Margin (95% CI)** | **Maximum probability (95% CI)** |
|---|---|---|
| Qwen, English | 0.734 (0.715 to 0.752) | 0.736 (0.717 to 0.754) |
| Qwen, Arabic | 0.646 (0.629 to 0.661) | 0.648 (0.631 to 0.663) |
| Jais, English | 0.600 (0.580 to 0.622) | 0.615 (0.595 to 0.637) |
| Jais, Arabic | 0.579 (0.561 to 0.595) | 0.595 (0.577 to 0.611) |

*Table C3. AUROC for predicting correctness from the top-two margin and from the maximum option probability; 95% percentile bootstrap intervals over questions (500 resamples).*